# A Zero-shot Learning Method Based on Large Language Models for Multi-modal Knowledge Graph Embedding

Bingchen Liu[1], Jingchen Li[1], Yuanyuan Fang[2], Xin Li[1*]

[1]School of Software, Shandong University, No. 1500 Shunhua Road, Jinan, 250000, Shandong Province, China.
[2]School of Computer and Artificial Intelligence, Shandong University of Finance and Economics, No. 7366, Second Ring East Road, Jinan, 250014, Shandong Province, China.

*Corresponding author(s). E-mail(s): lx@sdu.edu.cn;
Contributing authors: lbcraf2018@126.com;
202100300067@mail.sdu.edu.cn; 2896299017@qq.com;

**Abstract**

Zero-shot learning (ZL) is crucial for tasks involving unseen categories, suchas natural language processing, image classification, and cross-lingual transfer.Current applications often fail to accurately infer and handle new relations orentities involving unseen categories, severely limiting their scalability and prac-ticality in open-domain scenarios. ZL learning faces the challenge of effectivelytransferring semantic information of unseen categories in multi-modal knowledgegraph (MMKG) embedding representation learning. In this paper, we proposeZSLLM, a framework for zero-shot embedding learning of MMKGs using largelanguage models (LLMs). We leverage textual modality information of unseencategories as prompts to fully utilize the reasoning capabilities of LLMs, enablingsemantic information transfer across different modalities for unseen categories.Through model-based learning, the embedding representation of unseen cate-gories in MMKG is enhanced. Extensive experiments conducted on multiplereal-world datasets demonstrate the superiority of our approach compared tostate-of-the-art methods.

# 1 Introduction

Multi-modal knowledge graph (MMKG) is a kind of knowledge representation structure that integrates multiple modalities, such as text, images, and audio. It providesa more comprehensive description of entities and relations. MMKG embedding representation learning maps multi-modal information into a low-dimensional vectorspace, capturing the semantic features of entities and relations to support reasoningand knowledge expansion tasks. MMKGs have been applied in fields such as onlinesearch [1], Intelligent Transportation System [2], and medical data analysis [3]. How-ever, a large number of unseen categories involving new relations or entities frequentlyemerge across various domains, severely limiting the development and application ofdiverse systems in the era of big data.

Zero-shot learning (ZL) has been proposed to address this issue, with current meth-ods primarily focusing on establishing a mapping relation from seen classes to unseenclasses in multi-modal scenarios. For instance, a zero-shot learning framework [4] intro-duces a semantic knowledge graph (KG) to capture correlations between classes. Thissubstantially alleviates the severe limitations imposed by unseen categories involvingnew relations or entities on practical applications. However, in the context of ZL forembedding representations in current MMKGs, the semantic information of unseencategories often fails to be effectively transferred, resulting in significantly insufficientaccuracy.

In this work, we introduce ZSLLM, an innovative framework designed for multi-modal knowledge graph representation learning in zero-shot settings, which effectivelyharnesses the capabilities of large language models (LLMs). LLMs possess powerfulnatural language understanding and generation capabilities, enabling them to capturedeep semantic relations through rich context and demonstrate exceptional generaliza-tion and reasoning abilities across multiple tasks. We use the known information ofunseen classes as prompts, leveraging LLMs to assist in generating similar auxiliaryfeatures for unknown information, thereby enhancing semantic information transferamong unseen categories. By leveraging their strong natural language capabilities,LLMs can utilize textual information of unseen categories as prompts to achievecross-modal semantic alignment. Their outstanding reasoning ability and generaliza-tion further enable the effective capture of semantic features for unseen categories,addressing the challenges of ZL in MMKGs.

However, ZL requires models to reason about unseen categories, and while LLMspossess strong generalization capabilities, their knowledge may be too dispersed ornot optimized for specific tasks. Knowledge distillation (KL) can extract and transfergeneral knowledge (such as semantic relations and contextual understanding) fromLLMs to MMKG embedding tasks. Through distillation, the student model can bettercapture cross-modal associations and semantic information from the teacher model,thereby enhancing performance in zero-shot scenarios.

Our main contributions are as follows:

- To the best of our knowledge, it is the first work to apply LLMs to the zero-shot embedding representation learning scenario in MMKGs, demonstrating thesuperiority of LLMs in addressing this problem.

• Based on the proposed ZSLLM model, we achieve effective semantic informationtransfer for unseen categories in multi-modal knowledge graph scenarios, enablingzero-shot embedding representation learning.
• By leveraging knowledge distillation to fully capture cross-modal associationsand semantic information, the issue of LLMs potentially having overly dispersedknowledge or lacking optimization for specific tasks is effectively mitigated.
• Extensive experiments demonstrate that our proposed framework outperforms all existing baseline methods.

# 2 Related Works

Our work is closely connected to three key research areas: knowledge graph embed-dings, zero-shot learning, and large language models. These foundational domainscontribute significantly to our research framework, as elaborated below.

## 2.1 Knowledge graph embeddings

KG embeddings (KGEs) are vector representations of entities and relations in KGs,enabling them to be processed by machine learning models. Currently, KGE methodscan be classified into three main categories: translation-based models, bilinear-basedmodels, and neural network-based models. These models aim to minimize the distancebetween the head entity plus the relation vector and the tail entity vector.

### 2.1.1 Translation-based models

The fundamental idea behind translation-based models is that the embedding of thehead entity plus the embedding of the relation should approximate the embeddingof the tail entity. For instance, TransH [5] models a relation as a hyperplane andperforms a translation operation on it, thereby effectively preserving the relationalmapping properties. The form of this category of methods is relatively simple, but theparameters are often limited, which makes them unsuitable for the demands of somecomplex scenarios.

### 2.1.2 Bilinear-based models

In contrast, bilinear-based models employ a product-form scoring function to assessthe semantic relevance between entities and relations. By utilizing correlation as thecompositional operator, HOLE [6] is capable of capturing rich interactions while main-taining high computational efficiency, thereby facilitating the training process. Thiscategory of methods caters to the needs of more complex scenarios, but there is stillroom for improvement.

### 2.1.3 Neural network-based models

Neural network-based models input the triplet information from the knowledge graphinto neural networks to evaluate and score the triplets. Fan et al. [7] achieve KG embed-ding of full hyperplane through Lorentz transformation. Concept2Box [8] uses dual geometric representations to jointly embed the two views of KG. Xu et al. [9] introduce the GeomE method, which employs multivector representations and geometricproducts to effectively model entities and relations, subsequently scoring them. Caoet al. [10] introduce dual quaternions to score similarity. Zhao et al. [11] use a multi-plex relational graph attention network to learn on heterogeneous relational graph andgenerate embeddings. In contrast to the single-modal neural network-based modelsdiscussed, multi-modal neural network-based models have also been proposed in recentyears to address KGE scenarios that involve additional modalities such as images.As a dual fusion multi-modal knowledge graph embedding framework, DFMKE. [12]tackles entity alignment by combining the strengths of early and late fusion methods,offering a comprehensive fusion solution. Neural network-based models are power-ful for capturing non-linear patterns but require high computational resources andlarge datasets. Unlike traditional single-modal KGE methods, this paper exploresembedding representation learning for MMKG under unseen class data samples.

## 2.2 Zero-shot learning

Zero-shot learning is a machine learning paradigm that enables models to recognizeand classify unseen categories during inference. This is achieved by leveraging aux-iliary information, such as semantic attributes, textual descriptions, or pre-trainedembeddings, to bridge the gap between seen and unseen classes. Current approachesto ZL are generally categorized into two types: inductive ZL and transductive ZL.

### 2.2.1 Inductive ZL

These models rely solely on the training data without access to unseen class information. Liu et al. [13] propose POPRNet that incorporates discriminative partsemantics and object-centric semantics guided by semantic intensity to improve cross-domain transferability. Chen et al. [14] improve the ability to recognize unseenclasses by building explanatory graphs and partial aggregations to reduce virtual con-nections. PSVMA+ [15] handles attribute diversity and instance diversity throughcross-granularity learning and fusion of multi-granularity features.

### 2.2.2 Transductive ZL

These models incorporate information from the distribution of unseen class data during inference. Wu et al. [16] propose a prototype-augmented self-supervised generativenetwork that separates the generated unseen features and the generated seen featuresto solve bias problems. Chen et al. [17] propose a DSP method to align the empiri-cally predefined semantic prototypes and the real prototypes for class-related featuresynthesis. Bi-VAEGAN [18] enhances distribution alignment between the visual andauxiliary feature spaces, thereby effectively mitigating domain shift.

Inductive ZL generalizes to unseen classes based on learned features, offering flexibility but struggling with limited or dissimilar training data. Transductive ZL adaptsto specific test sets for higher accuracy but sacrifices generalization and depends ontest data distribution. The model proposed in this paper belongs to transductive ZL, where the training process leverages textual modality information of unseen classes toinfer the image distribution of unseen class data for training.

## 2.3 Large language models

LLMs are a type of advanced artificial intelligence system trained on massive amountsof text data to understand and generate human-like language. Currently, LLMs canbe categorized into two types: general-purpose models and specialized models.

### 2.3.1 General-purpose models

These models are designed for a wide range of tasks. Touvron et al. [19] train LLaMAusing publicly available datasets and the scaling laws. Bi et al. [20] propose DeepSeekby delving into the scaling law. Adler et al. [21] propose Nemotron by combining LLMand SFT and fine-tuning preferences.

### 2.3.2 Specialized models

These models are fine-tuned for specific applications or domains. Shen et al. [22]propose a model-agnostic framework for learning custom input tags, which are param-eterized as continuous vectors attached to the embedding layer of LLM to fine-tunethem. Dai et al. [23] propose the DeepSeekMo architecture, which improves specializa-tion by finely partitioning experts and isolating shared experts. Lee et al. [24] proposethe OWQ method to achieve LLMs fine-tuning with minimal memory overhead.

General-purpose models demonstrate versatility and adaptability across diversetasks, though they may not achieve optimal performance in specific domains. In con-trast, specialized models deliver superior accuracy in targeted tasks but exhibit limitedflexibility, often requiring retraining for adaptation to broader applications. The modelproposed in this paper is based on a general-purpose large model.

# 3 The proposed Framework

In this chapter, we first outline the fundamental process of developing the overallZSLLM framework. Subsequently, we provide a detailed description of each individualmodule within the proposed framework. The overall framework diagram we proposedis shown in Figure 1.

## 3.1 Framework outline

As shown in Figure 1, the input consists of two parts, the first part is the text featuresof seen and unseen classes, and the second part is the visual features of seen classes.We use the chatGPT interface to obtain word embeddings for the text features ofseen classes. For seen class image features, we use CNN [25] to convert them intoembedded features for input. In the model, we first input the image and text featuresof the seen class into the seen class learning module to learn the recognition experiencewithin the seen class. At the same time, we import the text features of unseen classesinto the unseen class recognition module to fully capture the available information of unseen classes. Finally, based on the learning experience of the first two modules, weuse the unseen class classification module to classify images of unseen classes. Theultimate goal of the proposed model is to classify images of unseen classes, such as theclassification of images of horses and cats shown in Figure 1.

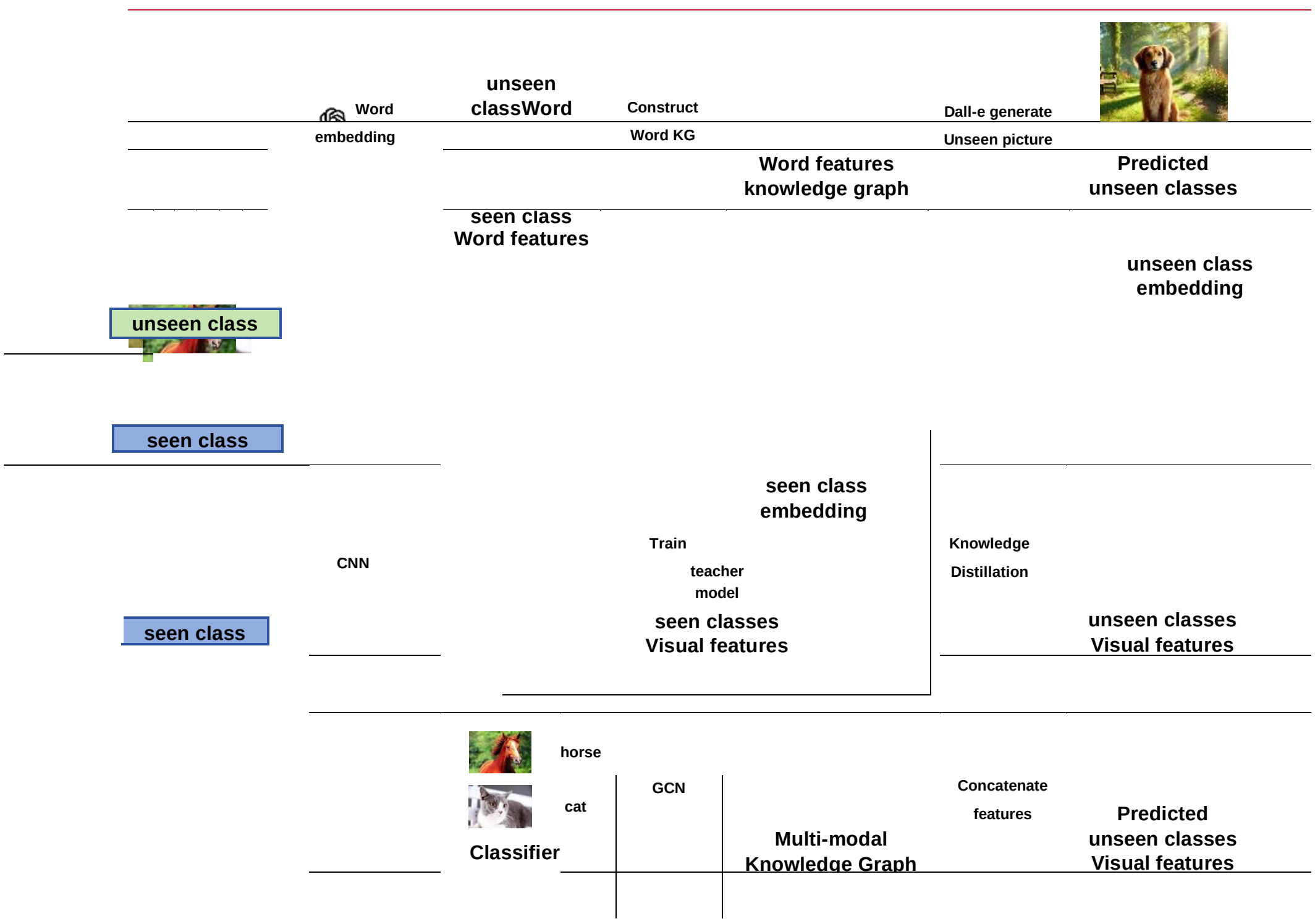

**Fig. 1** Overall framework of ZSLLM

## 3.2 Seen class learning module

In order to fully solve the problem of difficult knowledge generalization caused bythe lack of target samples in zero sample learning, a knowledge distillation model isintroduced in ZSLLM. We train the teacher model in the knowledge distillation modelbased on seen class text features and visual features to facilitate the generalization ofknowledge to classification problems of unseen classes.

Generate word features based on chatGPT interface for seen class text labels, andgenerate visual features based on CNN for seen class image features:

$$x = \mathrm{concat}(Ws, V s) \tag{1}$$

Here, $Ws$ represents the word features of seen - class text labels generated via theChatGPT interface, $V s$ represents the visual features of seen - class image featuresgenerated by CNN, and $x$ is the concatenated feature vector of $Ws$ and $V s$.

$$x = \mathrm{reshape}(x, [\text{batch size}, 1, 28, 28]) \tag{2}$$

In this equation, the feature vector $x$ is reshaped into a 4 - D tensor with dimensions[batch size, 1, 28, 28].

$$x1 = \mathrm{ReLU}(\mathrm{Conv}(x, W1) + b1) \quad (3)$$

Here, $W1$ is the convolution kernel, $b1$ is the bias term, and $x1$ is the output obtainedafter performing a convolution operation on $x$ with $W1$, adding the bias $b1$, and thenapplying the ReLU activation function.

$$x2 = \mathrm{Pool}(x1) \quad (4)$$

In this formula, $x2$ is the result of applying a pooling operation (e.g., max-pooling oraverage-pooling) on $x1$.

$$x3 = \mathrm{flatten}(x2) \quad (5)$$

This equation flattens the multi-dimensional tensor $x2$ into a one-dimensional vector$x3$.

$x4 = \mathrm{ReLU}(W4 \cdot x3 + b4)$(6)Here, $W4$ is a weight matrix, $b4$ is a bias term. $x4$ is the output after a linear trans-formation of $x3$ by $W4$, adding the bias $b4$, and then applying the ReLU activationfunction.

$\mathrm{output} = W5 \cdot x4 + b5$(7)In this final equation, $W5$ is a weight matrix, $b5$ is a bias term, and output is the finaloutput of the entire model, which is a linear transformation of $x4$ by $W5$ with theaddition of $b5$. Then, we train the teacher model based on and:

$$x = \mathrm{concat}(Ws, Vs) \quad (8)$$

Here, $x$ is the result of concatenating the vectors $Ws$ and $Vs$.

$x_1 = \mathrm{ReLU}\,(\mathrm{Dropout}(W_1 \cdot x + b_1))$(9)In this formula, $x_1$ is the output after applying ReLU and Dropout to the linearcombination $W_1 \cdot x + b_1$, where $W_1$ is a weight matrix, $b_1$ is a bias vector, and $x$ is theinput.

$x_2 = \mathrm{ReLU}\,(\mathrm{Dropout}(W_2 \cdot x_1 + b_2))$(10)For this equation, $x_2$ is obtained by applying ReLU and Dropout to $W_2 \cdot x_1 + b_2$, with$W_2$ as a weight matrix, $b_2$ as a bias vector, and $x_1$ as the input from the previous layer.

$x_3 = W_3 \cdot x_2 + b_3$(11)Finally, $x_3$ is the output of the linear transformation $W_3{\cdot}x_2{+}b_3$, where $W_3$ is a weightmatrix, $b_3$ is a bias vector, and $x_2$ is the input.

Then, based on knowledge distillation, we transfer the knowledge from the

$$x = \mathrm{concat}(Ws, Vs) \quad (12)$$

This equation shows that the vector $x$ is obtained by concatenating the feature vectors $Ws$ and $Vs$.

$$\text{teacher preds} = \text{teachermodel}(x) \tag{13}$$

This equation indicates that teacher preds are the predictions generated by the teachermodel when taking $x$ as input.

$$\text{student preds} = \text{studentmodel}(x) \tag{14}$$

This equation implies that student preds are the predictions produced by the studentmodel with $x$ as the input.

Here we optimize based on the following formula:

$$\text{student loss} = \text{CrossEntropyLoss}(\text{student preds}, \text{target}) \tag{15}$$

This equation calculates the student loss using the cross - entropy loss function withstudent preds and target as inputs.

$$\text{distillation loss} = \text{KLDivLoss}\left(\log(\text{softmax}(\text{student_preds}/\text{temp})), \text{temp}\right) \quad \text{softmax}(\text{teacher preds}/\text{temp}) \quad \text{temp} \tag{16}$$

This equation computes the distillation loss by applying the Kullback - Leibler divergence loss function on the softened log - probabilities of student preds andteacher preds with a temperature temp.

$$\text{loss} = \alpha \cdot \text{student loss} + (1 - \alpha) \cdot \text{distillation loss} \tag{17}$$

This equation calculates thetotallossbycombiningthe student loss and distillation loss with a weighted sum using the hyperparameter $\alpha$.

### 3.3 Unseen class recognition module

In order to obtain auxiliary recognition information for unseen classes, we generateword features for unseen classes based on the chatGPT interface. Then, based on theknowledge graph embedding model [4], we import the structural information of themultimodal knowledge graph into:

$$\hat{F}_{s+u} = \sigma(D^{-1}A\sigma(D^{-1}AF_{s+u}\theta)\theta) \tag{18}$$

In this formula, $\hat{F}_{s+u}$ is the updated feature representation for the combination of seennon - linearity into the model, such as the sigmoid or ReLU function. $D$ is the degreematrix of the graph, which is a diagonal matrix where each diagonal element representsthe degree of a corresponding node in the graph. $A$ is the adjacency matrix of thegraph, which describes the connections between nodes. $F_{s+u}$ is the original featurematrix for the combination of seen and unseen classes, and $\theta$ is a learnable parametermatrix used to transform the features.

$$F_{s+u} = GDAN(w_{u\cdot} c_{c\cdot}) \tag{19}$$

Here, $F_{s+u}$ represents the feature matrix for the combination of seen and unseen classes. $G$ is a graph-related transformation function that is designed to capture thegraph-based structure and features. $DAN$ stands for a deep averaging network, whichis a type of neural network used for text processing. $w_{u\cdot}$ denotes the word features asso-ciated with the unseen classes, and $c_{c\cdot}$ is a matrix related to class - specific information,which may contain semantic or statistical information about different classes.

Fully tapping into the massive pre training knowledge of LLMs in ZSLLM providesa broad knowledge foundation for zero sample learning. Using Dall-e to generate sim-ilar invisible class image features for student models to train and learn, we generatereference feature vectors based on CNN for the generated images:

$$V''_g = \text{softmax}(V_g/T) \tag{20}$$

In this formula, $V''_g$ is the output of applying the softmax function to the scaled $V_g$.softmax is the softmax activation function that converts a vector of real numbersinto a probability distribution. $V_g$ is the original generated feature vector, and $T$ is atemperature parameter used to control the smoothness of the softmax output.

$$L = \alpha \cdot \text{Hard}_{Loss} + (1 - \alpha) \cdot \text{Soft}_{Loss} \tag{21}$$

In this equation, $L$ represents the overall loss function. $\alpha$ is a hyperparameter thatcontrols the trade - off between $\text{Hard}_{Loss}$ and $\text{Soft}_{Loss}$. $\text{Hard}_{Loss}$ is a hard loss function,and $\text{Soft}_{Loss}$ is a soft loss function.

$$L = \alpha \cdot \text{CE}(V_g, V_c) + (1 - \alpha) \cdot \text{CE}(V''_g, V_c) \tag{22}$$

Here, $L$ is still the overall loss. $\alpha$ has the same meaning as before. CE denotes thecross - entropy loss function. $V_g$ is the generated feature vector, $V_c$ is the target featurevector, and $V''_g$ is a modified version of $V_g$.

## 3.4 Unseen class classification module

In this module, we first aggregate the seen and unseen text and visual feature vectorsobtained from the first two modules based on a multimodal knowledge graph

$$\mathbf{M}_c = \text{Concat}(\text{MultiModal}_c, \mathbf{E}_c) \tag{23}$$

Then, we use GCN to determine the specific category to which unseen class images belong:

$$\tilde{A} = A + I \tag{24}$$

Here, $\tilde{A}$ is the adjacency matrix with added self-loops, $A$ is the original adjacencymatrix, and $I$ is the identity matrix.

$$\tilde{D}_{ii} = \sum_j \tilde{A}_{ij} \tag{25}$$

Here, $\tilde{D}$ is the degree matrix of $\tilde{A}$, where $\tilde{D}_{ii}$ is the sum of the $i$-th row of $\tilde{A}$.

$$H^{(0)} = \mathbf{M}_c \tag{26}$$

Here, $H^{(0)}$ is the initial feature matrix, and $\mathbf{M}_c$ represents the feature matrix of theunseen classes.

$$H^{(1)} = \text{ReLU}(\tilde{D}^{-\frac{1}{2}}\tilde{A}\tilde{D}^{-\frac{1}{2}}H^{(0)}W^{(0)}) \tag{27}$$

Here, $H^{(1)}$ is the output of the first GCN layer, $\tilde{D}^{-\frac{1}{2}}\tilde{A}\tilde{D}^{-\frac{1}{2}}$ is the normalized adjacencymatrix, and $W^{(0)}$ is the weight matrix for the first layer.

$$\text{Classification results} = \text{Softmax}(H^{(1)}) \tag{28}$$

Here, the classification results are obtained by applying the Softmax function to $H^{(1)}$.

# 4 EXPERIMENTS

In this section, we introduce our comparison results with other existing models.In addition, we also validate the effectiveness of various components and partialparameters of the model.

## 4.1 Experimental setting

### 4.1.1 Dataset

Our main experiment is tested on three datasets, namely ImageNet [26], AWA2 [27],and Attribute Pascal and Yahoo (aPY) [28].

ImageNet: This dataset aims to use an average of 500-1k clean and high-resolutionimages to form the majority of WordNet's [29] 80k synonyms. Additionally, it aimsto construct a comprehensive semantic tree comprising millions of ImageNet images,structured according to the WordNet framework. ImageNet currently comprises 12hierarchical categories, 5,247 synsets, and 3.2 million annotated images. The ImageNetdataset surpasses many existing datasets in terms of scale, diversity, and annotationaccuracy.

AWA2: This dataset collects 37322 images from 50 categories of the AWA1 dataset[27] from public network resources such as Flickr and Wikipedia. All AWA2 imagescan be used for free. In addition, the images in this dataset cannot overlap with theimages in the AWA1 dataset. The AWA2 dataset contains 50 animal categories. Eachclass contains an average of 746 images, with the mole class being the least numerousclass with 100 images. Horses are the class with the highest number of images, with1645 images.

Attribute Pascal and Yahoo (aPY): This data set is a coarse-grained small-scaledataset with 64 attributes. Use 20 classes in training and 12 classes in testing, for atotal of 32 classes. The dataset of primitive animals with attributes (AWA1) [30] hasa coarse granularity and moderate scale.

**Algorithm 1** Zero-Shot Learning Model Algorithm

**Input**:
- $Ws$: Text features of seen classes
- $Vs$: Visual features of seen classes
- $Wu$: Text features of unseen classes
- target: True labels of seen classes
- $\alpha$: Hyper-parameter for loss combination
- temp: Temperature parameter for distillation loss

**Output**: Classification results of unseen class images
**// Seen class learning module**
Calculate $x$ according to Eq. 1
Reshape $x$ according to Eq. 2
Calculate $x1$ according to Eq. 3
Calculate $x2$ according to Eq. 4
Calculate $x3$ according to Eq. 5
Calculate $x4$ according to Eq. 6
Calculate output according to Eq. 7
Calculate $x$ according to Eq. 8
Calculate $x_1$ according to Eq. 9
Calculate $x_2$ according to Eq. 10
Calculate teacher preds according to Eq. 11
Calculate $x$ according to Eq. 12
Calculate teacher preds according to Eq. 13
Calculate student preds according to Eq. 14
Calculate student loss according to Eq. 15
Calculate distillation loss according to Eq. 16
Calculate loss according to Eq. 17
Update student model parameters based on loss
**// Unseen class recognition module**
Generate word features for unseen classes based on ChatGPT interface
Calculate $F_{s+u}$ according to Eq. 18
Generate similar invisible class image features using Dall-e
Generate reference feature vectors $V_g$ based on CNN for the generated images
Calculate $V''_g$ according to Eq. 20
Calculate $L$ according to Eq. 22
**// Unseen class classification module**
Aggregate seen and unseen text and visual feature vectors:
Calculate $\mathbf{M}_c$ according to Eq. 23
**// Apply GCN for classification**
Let $A$ be the adjacency matrix of the graph representing the relations between features
Calculate $\tilde{A}$ according to Eq. 24
Calculate $\tilde{D}$ according to Eq. 25
Set $H^{(0)}$ according to Eq. 26
Let $W^{(0)}$ be the learnable weight matrix of the first GCN layer
Calculate $H^{(1)}$ according to Eq. 27
Calculate Classification results according to Eq. 28 **return** Classification results

### 4.1.2 Assessment measures

We use Hits@n as our experimental evaluation metric. Hits@n The average proportionof knowledge in the prediction process is lower than n. In the experiment, the higherthe Hits@n value, the better the experimental effect. Hits@n The calculation methodis as follows:

$$\text{Hits@n} = 1/|S| \cdot \sum_{i=1}^{|S|} indicator(\text{rank}_i \leq n)$$

, where the function of indicator is: if the condition is true, the function value is 1;otherwise, it is 0. $|S|$ is the number of quadruple sets, and $\text{rank}_i$ refers to the linkprediction ranking of the $i$-th quadruple.

### 4.1.3 Set details

For the GCN used, in order to avoid overfitting, we refer to [4] and use the dropoutfunction [31], with a dropout rate of 0.5 for each layer. The Adam optimizer is trainedin 3000 rounds, with a learning rate set to 0.001 and weight decay of 0.0005.

**Table 1** Top-K Accuracy of All Methods on ImageNet Dataset

| | 2-hops | | | | | 3-hops | | | | | All | | | | |
|---|---|---|---|---|---|---|---|---|---|---|---|---|---|---|---|
| | 1 | 2 | 5 | 10 | 20 | 1 | 2 | 5 | 10 | 20 | 1 | 2 | 5 | 10 | 20 |
| ConSE | 8.3 | 12.9 | 21.8 | 30.9 | 41.7 | 2.6 | 4.1 | 7.3 | 11.1 | 16.4 | 1.3 | 2.1 | 3.8 | 5.8 | 8.7 |
| SYNC | 10.5 | 17.7 | 28.6 | 40.1 | 52 | 2.9 | 4.9 | 9.2 | 14.2 | 20.9 | 1.4 | 2.4 | 4.5 | 7.1 | 10.9 |
| EXEM† | 12.5 | 19.5 | 32.3 | 43.7 | 55.2 | 3.6 | 5.9 | 10.7 | 16.1 | 23.1 | 1.8 | 2.9 | 5.3 | 8.2 | 12.2 |
| GCNZ♯ | 19.8 | 33.3 | 53.2 | 65.4 | 74.6 | 4.1 | 7.5 | 14.2 | 20.2 | 27.7 | 1.8 | 3.3 | 6.3 | 9.1 | 12.7 |
| SGCN | 25.51 | 39.4 | 58.88 | 70.73 | 80.03 | 5.79 | 10.3 | 18.39 | 26.41 | 35.94 | 2.71 | 4.74 | 8.87 | 13.15 | 18.78 |
| DGP | 26.16 | 39.9 | 59.31 | 71.14 | 80.29 | 6.13 | 10.37 | 18.78 | 27.06 | 36.83 | 2.87 | 4.91 | 9.08 | 13.54 | 19.32 |
| our | 33.46 | 48.32 | 62.87 | 75.41 | 84.26 | 8.47 | 13.84 | 23.05 | 32.6 | 46.58 | 4.46 | 8.73 | 13.1 | 18.54 | 28.74 |

**Table 2** Top-1 Accuracy Results for Unseen-Classes on AWA2 and APY.

| | Implicit Correlation Methods | | | | | Explicit Correlation Methods | | | |
|---|---|---|---|---|---|---|---|---|---|
| | ConSE | Devise | SYNC | SE - GZSL | Gaussian - Ort | GCNZ | SGCN | DGP | Our |
| AWA2 | 44.5 | 59.7 | 46.6 | 69.2 | 70.5 | 70.7 | 76.64 | 77.3 | 85.32 |
| aPY | 26.9 | 39.8 | 23.9 | 45.3 | 17.91 | 19.03 | 19.08 | 26.87 | 58.4 |

## 4.2 Main experiments

### 4.2.1 Comparison method

ConSE [32] maps images to a semantic embedding space through convex combinationsof class label embedding vectors, without the need for additional training. EXEM [33]utilizes clustering structures in the semantic embedding space to enable semantic rep-resentations to predict the position of their corresponding visual examples, thereby performing structural constraints. SYNC [34] aligns the semantic space from exter-nal information with the model space that focuses on recognizing visual features.Devise [35] proposes using trained word embeddings instead of one hot vectors as thelabels to be predicted by the model, and then fine tune based on this. SE-GZSL [36]can generate novel examples from visible/invisible classes, given their respective classattributes, and these examples can then be used to train any ready-made classificationmodel. Gaussian Ort [37] sets a simple learning objective inspired by linear discrimi-nant analysis, kernel target alignment, and kernel polarization methods. GCNZ [38] isbased on graph convolutional networks and uses both semantic embedding and clas-sification relationships to predict classifiers. SGCN [39] designs direct links betweendistant nodes based on a dense graph propagation module. DGP [39] allows the hier-archical graph structure of KGs to be utilized through additional connections. A zeroshot learning method based on a multi modal KG [4] introduces a semantic KG tocapture the correlations between classes and learn the visual feature representationsof all classes through this correlation. It can be seen that our model outperforms allbaseline models and maintains optimal performance on all three datasets.

### 4.2.2 Result of the comparison

In Table I and Table II, we present the experimental results of the main experiment. Itcan be seen that we outperform existing zero sample learning prediction models in allthree datasets, achieving optimal performance in multimodal KG classification tasks.Unlike traditional prediction models, our proposed model fully utilizes the massivepre-training knowledge of LLMs, providing extensive foundational knowledge for zerosample learning and enhancing the model's generalization. In addition, based on theknowledge distillation model, the ability to understand new categories of data hasbeen improved. As a result, our model sustains the most competitive task performancethat is currently attainable.

## 4.3 Ablation analysis

To verify the effectiveness of the LLM and the knowledge distillation componentsin our proposed model, we perform ablation experiments. The experimental resultsshow that removing each component leads to a decrease in performance, proving theimportance of each component in the model.

Specifically, after removing the visible class component, the experimental resultdrop to 3.24%, a decrease of 1.22% compared to the main experiment result of 4.46%.This fully demonstrates the significant role of the seen class component in improvingthe model's performance. Similarly, after removing the knowledge distillation com-ponent, the experimental result drop to 3.97%, also lower than the main experimentresult of 4.46%, which validates that the knowledge distillation model can effectivelyleverage pre-trained knowledge, helping the model to perform better inference andprediction in zero-shot learning problems.

### 4.4 Hyper-parameter Analysis

To verify the effectiveness of the parameters in our proposed model, we conduct twosets of parameter variation experiments.

#### 4.4.1 Distillation temperature

We present the experimental results of the distillation temperature variation in Figure2. We vary the distillation temperature from 2.0 to 5.0, with a magnitude of 0.5 foreach change. It can be seen that as the distillation temperature changes from 2.0 to 5.0,the accuracy of the experiment gradually increases first, reaching its highest value at3.0, and then decreases. This indicates that as the distillation temperature increases,more inter category relation information can be provided during the knowledge distil-lation process. However, after reaching a certain distillation temperature, the effectiveinformation that distinguishes the original categories can be lost. Therefore, in thezero-sample learning process, richer empirical information can be utilized to match andinfer unseen categories, which is beneficial for improving accuracy. However, attentionshould be paid to increasing the range of distillation temperatures.

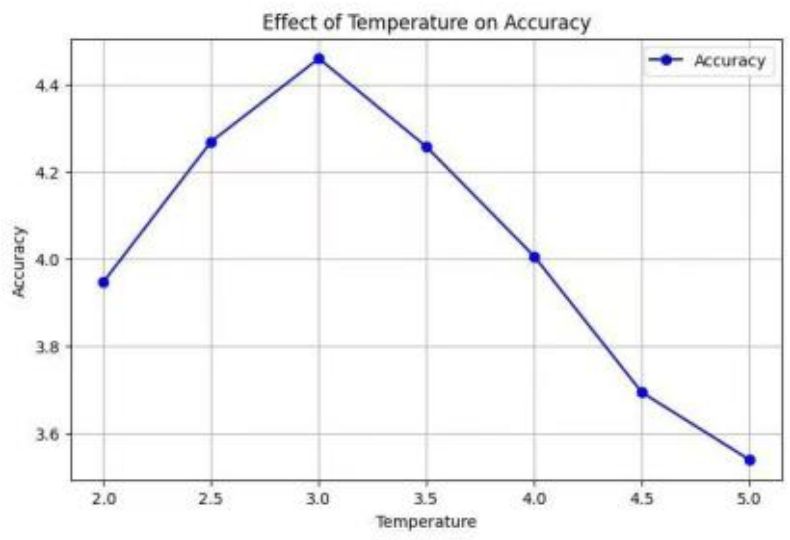

**Fig. 2** Experimental results of distillation temperature variation from 2.0 to 5.0. The accuracyinitially increases, peaks at 3.0, and then decreases, indicating the optimal temperature for knowledgedistillation.

#### 4.4.2 GCN layers

We present the experimental results of GCN layer changes in Figure 3. We vary thenumber of GCN layers from 1000 to 5000, with each change being 500. It can be seenthat as the number of GCN layers increases, the accuracy of the experiment graduallyincreases and then decreases, reaching its highest value at 3000 layers. This is becauseas the number of layers in GCN increases, it can extract more complex features instructures such as KG. However, after reaching a certain number of layers, there canbe an issue of over smoothing, causing node features to become too similar. In thezero sample learning process, based on richer features, it is possible to infer unseencategories, thereby improving the accuracy of prediction. Therefore, it is possible tomoderately increase the number of GCN layers while avoiding overfitting.

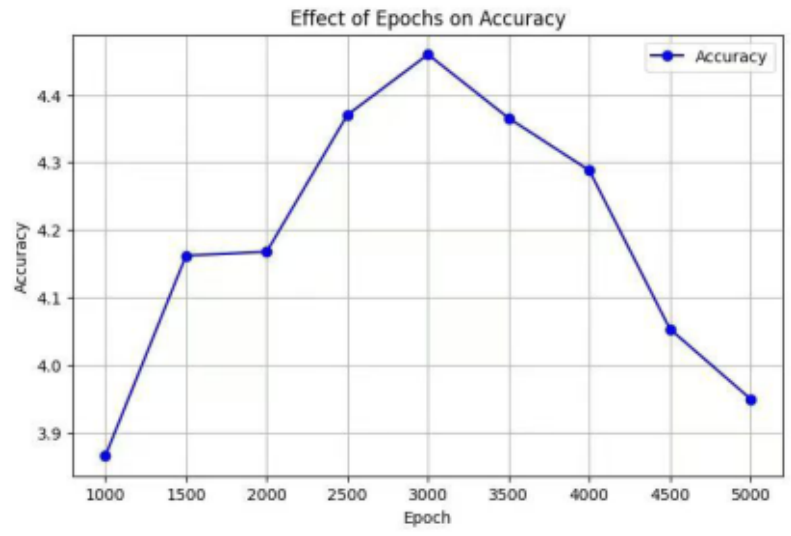

**Fig. 3** Experimental results of GCN layer changes from 1000 to 5000. The accuracy increases withthe number of layers, peaking at 3000 layers before decreasing, indicating the optimal number oflayers for feature extraction without overfitting.

## 5 Conclusion

The challenge of ineffective semantic information transfer for unseen categories inMMKG embedding representation learning has not been adequately addressed byprevious methods. In this paper, we propose a novel zero-shot learning framework,ZSLLM, which effectively enhances the representation learning of unseen categoriesin MMKG by establishing semantic connections across different modalities for theseunseen categories. To demonstrate the effectiveness of ZSLLM, we conduct extensiveexperiments on three real-world datasets. The experimental results show that ourframework outperforms all baselines. In future research, we will further explore theintegration of LLMs with other KG scenarios.

## 6 Acknowledgments

This research is sponsored by the Key R&D Program of Shandong Province,China (No.2023CXGC010801) and the National Key R&D Program of China(2019YFA0709401).

## Author Biographies

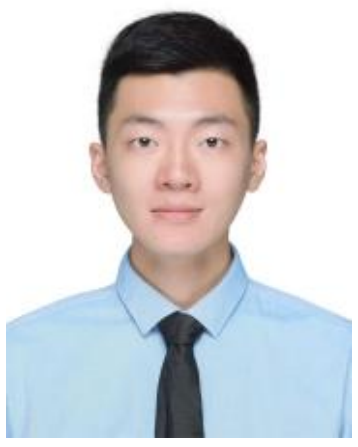

**Bingchen Liu** received the B.S. degree from the School of Com-puter Science and Technology, Shandong University of Finance andEconomics, Jinan, China, in 2022. He is currently pursuing the Ph.D.degree with the School of Software, Shandong University, Jinan.His research interests include knowledge graph alignment and graphlearning.

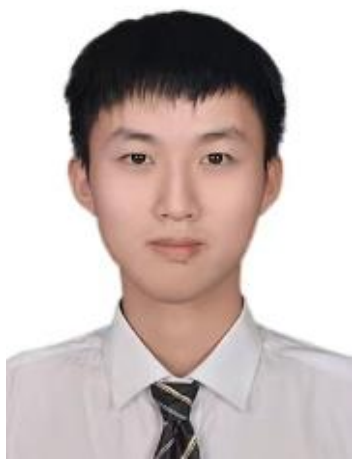

**Jingchen Li** is currently pursuing a Bachelor's degree in Data Sci-ence and Big Data Technology at Shandong University, Jinan, China.His research interests focus on large-scale language models, their opti-mization in multimodal applications, and data privacy challenges,aiming to advance innovation and practice in artificial intelligence.

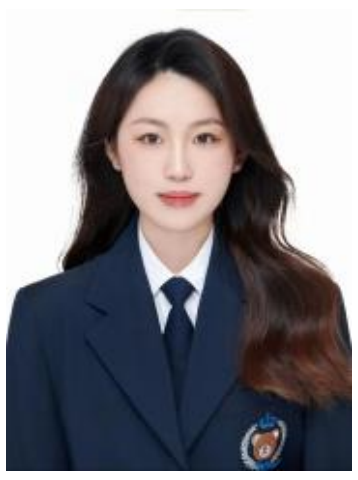

**Yuanyuan Fang** is currently pursuing the B.S. degree with theSchool of Computer Science and Technology, Shandong University ofFinance and Economics, Jinan, China. Her research interests includeimage processing and deep learning.

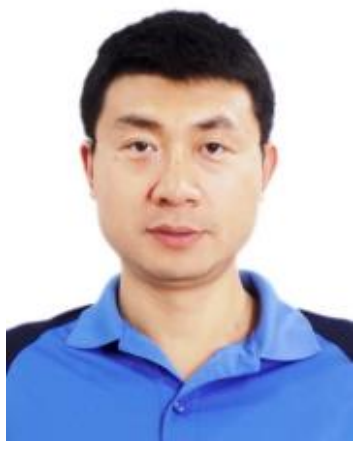

**Xin Li** received the Ph.D. degree from the Institute of Software,Chinese Academy of Sciences, Beijing, China, in 2008. He is currentlyan associate professor in the School of Software, Shandong University.He has authored or coauthored a number of papers including TCAD,DATE and ASP-DAC. His research interests cover knowledge graph,object detection, and heterogeneous computing systems.